# Exploring LLM Multi-Agents for ICD Coding

Rumeng Li[1,2], MS; Xun Wang[2], PhD; Hong Yu[1,2,3,4], PhD

[1]Manning College of Information & Computer Sciences, University of Massachusetts Amherst, Amherst, MA, USA;

[2]Center for Healthcare Organization & Implementation Research, VA Bedford Health Care System, Bedford, MA, USA;

[3]Center of Biomedical and Health Research in Data Sciences, University of Massachusetts Lowell, Lowell, MA, USA;

[4]Miner School of Computer & Information Sciences, University of Massachusetts Lowell, Lowell, MA, USA;

Corresponding Author:

Hong Yu, PhD

Director, Center of Biomedical and Health Research in Data Sciences

Miner School of Computer & Information Sciences, University of Massachusetts Lowell

1 University Avenue, Lowell, MA, 01854, United States

Phone: 1 9789343620

Fax: 1 9789343551

Email: hong_yu@uml.edu

**Abstract**

**Objective:** To address the limitations of Large Language Models (LLMs) in the International Classification of Diseases (ICD) coding task, where they often produce inaccurate and incomplete prediction results due to the high-dimensional and skewed distribution of the ICD codes, and often lack interpretability and reliability as well.

**Materials and Methods:** We introduce an innovative multi-agent approach for ICD coding which mimics the ICD coding assignment procedure in real-world settings, comprising five distinct agents: the patient, physician, coder, reviewer, and adjuster. Each agent utilizes an LLM-based model tailored to their specific role within the coding process. We also integrate the system with Electronic Health Record (HER)'s SOAP (subjective, objective, assessment and plan) structure to boost the performances. We compare our method with a system of agents designed solely by LLMs and other strong baselines and evaluate it using the Medical Information Mart for Intensive Care III (MIMIC-III) dataset.

**Results and Discussion:** Our multi-agent coding framework significantly outperforms Zero-shot Chain of Thought (CoT) prompting and self-consistency with CoT (CoT-SC) in coding common and rare ICD codes. An ablation study validates the effectiveness of the designated agent roles. it also outperforms the LLM-designed agent system. Moreover, our method achieves comparable results to state-of-the-art ICD coding methods that require extensive pre-training or fine-tuning, and outperforms them in rare code accuracy, and explainability. Additionally, we demonstrate the method's practical applicability by presenting its performance in scenarios not limited by the common or rare ICD code constraints.

**Conclusion:** The proposed multi-agent method for ICD coding effectively mimics the real-world coding process and improves performance on both common and rare codes. It also provides a level of interpretability and reliability that matches current leading methods, without the need for extensive pre-training or fine-tuning.

**Keywords:** Large language models, ICD-coding, Multi-agent

**Introduction**

The International Classification of Diseases (ICD) is a standardized system of codes that represent various clinical activities, such as diagnoses, procedures, and mortality causes. The practice of ICD coding, which involves assigning these standardized codes to clinical documentation, is pivotal for various healthcare operations such as billing, epidemiological studies, and enhancing care quality. An accurate coding process ensures the precise and uniform conveyance of clinical data, facilitating effective communication across the healthcare continuum[1].

Despite its significance, ICD coding presents a significant challenge due to the necessity of meticulously analyzing diverse and intricate clinical narratives to select appropriate codes from an extensive, hierarchical set. The current iteration, ICD-10, encompasses 68,000 diagnostic and 87,000 procedural codes, adding to the complexity of the task. Clinical notes, meanwhile, vary significantly in terms of length, format, content, and style. The Medical Information Mart for Intensive Care III (MIMIC-III) dataset [2], a widely used medical database, which is also used in this work, has clinical notes that are in length from less than 500 words to more than 3000 words. These pose significant challenges for both humans and machines in understanding clinical notes and assigning ICD codes correctly. Moreover, ICD coding deals with a large label space with a long-tail issue. In the MIMIC-III dataset, the top 10% of all ICD codes cover 85% of all code occurrences, while about 22% of codes occur no more than twice [3]. In addition, ICD coding is often subject to errors and inconsistencies, as different coders may interpret and apply the coding rules and criteria differently, or miss some relevant codes.

To tackle these challenges, automatic ICD coding has emerged as an important and promising research topic in natural language processing (NLP). Automatic ICD coding aims to develop machine learning methods that can automatically assign ICD codes to clinical notes, based on the natural language understanding and generation capabilities of the models [4,5]. Large language models (LLMs), such as GPT-* models [6] , Llama [7], Gemini [8] etc., have been explored for this purpose. However, recent studies on ICD coding with LLMs reveal the task's challenges, such as LLMs' lack of domain-specific knowledge and vocabulary, the multi-label and long-tailed code space, and the vulnerability to noisy or adversarial inputs [9–11]. Study has shown that, models like GPT-4, have relatively high recalls but low precisions in medical coding tasks [11], which limits their practical use. The work by [5] also found that LLMs evaluated were inadequate for medical code querying tasks due to the unsatisfying performances. Consequently, it is suggested that LLMs should not be deployed for medical coding purposes until further investigations are conducted.

To fully explore LLMs' potential in ICD coding, we propose a multi-agent system for this task and explore its performances under different settings. Multi-agent systems, as one of the recent advances, have shown their advantages in employing LLMs to create "interactive simulacra" that replicate human behaviors [12] to resolve complex tasks. Our multi-agent framework for ICD coding features five specialized agents (patient, physician, coder, reviewer and adjuster) using LLMs for role-specific tasks. The system is designed for dynamic interaction, strategic collaboration, and competition, enhancing the coding process's explainability and robustness. Figure 1 illustrates this interplay and the agents' strategic functions for optimal coding accuracy.

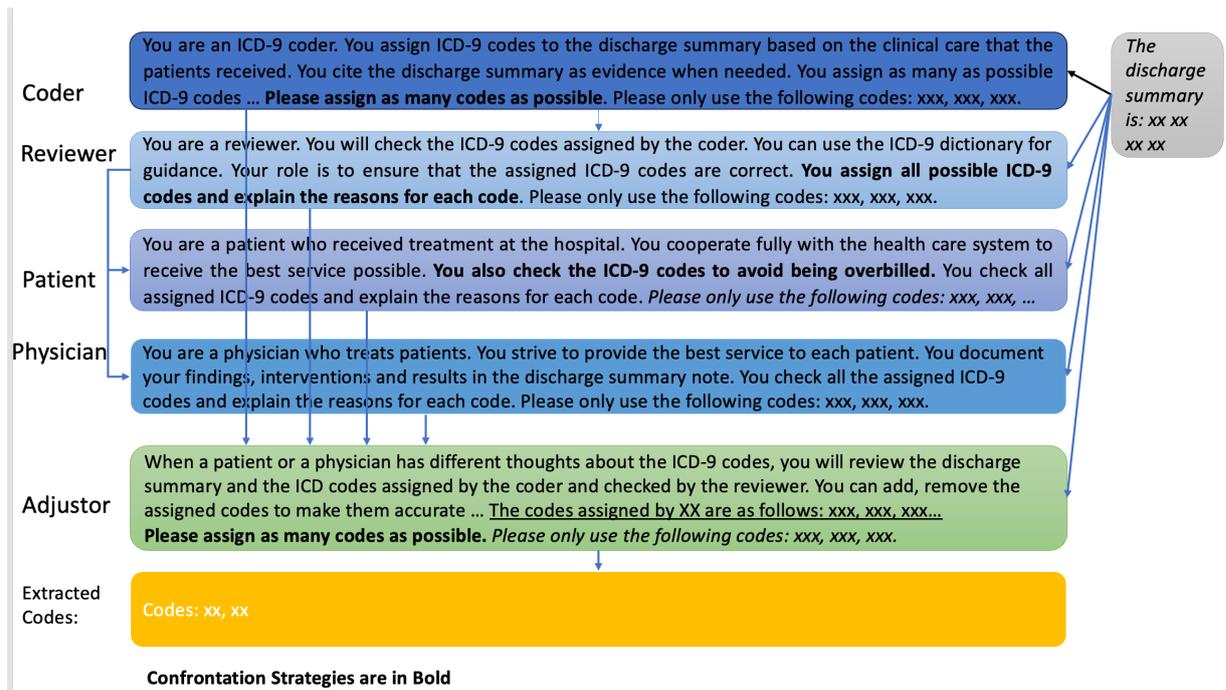

Figure 1: The Assignment of ICD Codes with Multi-agents. The workflow began when a patient came to the physician for help. The physician documented the clinical activities in the note. The coder assigned codes according to the note. The reviewer reviewed the codes. The patient and the physician reviewed the codes and asked the adjustor to review again if they disagreed with the codes. The adjustor made the final decisions. Confrontation Strategies are in Bold.

To further enhance the proposed system's performances, we leverage the SOAP (Subjective, Objective, Assessment, and Plan) structure of electronic health records (EHRs). Instead of directly asking LLMs to output the ICD codes, we feed them the subjective and objective sections, and prompt them to generate the assessment and plan sections. Then we instruct them to compare their own generation with the gold standard, and self-correct any hallucinations. Finally, we output the ICD codes. We designed multi-agents to accomplish the whole process. We observe that this reasoning and self-correction process improves the ICD prediction performance.

We also explored different methods to boost model performances. Existing research has shown that LLMs can achieve better results with integration of external tools or knowledge [13]. We observed an increase in results with the help of code description extracted from external resources. Similar gains were also observed when we adopted the confrontation strategies to make agents compete with each other.

The main contributions of our paper are as follows:

- We introduce a novel multi-agent system for automated ICD coding that mirrors the intricate and dynamic interplay between patients, physicians, and coders in the actual coding environment.
- We leverage the SOAP structure of EHR notes to enable LLMs self-correct hallucinations, boosting performance. A confrontational strategy among agents, coupled with the integration of external code knowledge, further refines our system's efficiency.

- Our methods substantially improve LLM's ICD coding performance over Chain of thought (CoT) prompting and self-consistency CoT (CoT-SC) prompting, achieving comparable performance with state-of-art ICD coding models requiring pre-training or fine-tuning. The ablation study validates the efficacy of the proposed agent roles and the effectiveness of our proposed strategy. We analyze the results and discuss the insights and limitations of our method and the existing methods, and provide some directions and suggestions for future work and improvement.

**Related Work**

In this section, we discuss the previous work on ICD coding and multi-agent systems that inspires this study.

**ICD Coding** ICD coding predicts expert labeled ICD codes from EHR notes using NLP models. Previous methods can be categorized as follows: Rule-based methods use hand-crafted rules to assign codes [14]. Methods like bag-of-words (BoW), word2vec (W2V) [15], or pre-trained embedding models are used to represent the notes and the codes [16,17] and map them into vectoral spaces for code assignment. Neural models like CNNs are widely used to encode and classify the notes and the codes [18,19]. Some methods propose to capture the sequential information with attentive LSTM or tree-of-sequences LSTM [20,21]. Some methods further enhance the neural models with attention mechanisms, dense connections, residual connections, or dilated convolutions [22–24]. Fine-tuning methods which use pre-trained language models (PLMs) that are fine-tuned on the medical-specific and special language of clinical note are widely adopted because of their ability to process domain-specific text [25]. Knowledge injection methods which use medical domain-specific knowledge such as hierarchy and synonym to improve ICD coding are also developed [4,26–30]. Active learning methods using extra human labeling are also studied to address the rare code issue[31].

In recent years LLMs have significantly enhanced the ability of machines to understand and generate natural language. LLM-based methods for ICD coding use LLMs such as GPT or T5 etc. to encode and generate the clinical notes and the ICD codes. [5] assesses GPT-3.5 and GPT-4 in generating ICD Codes and concludes that while the models appear to exhibit a general conceptual understanding of the codes and their descriptions, they have a propensity for hallucinating key details, suggesting underlying technological limitations of the base LLMs. They suggest a need for more rigorous LLM augmentation strategies and validation prior to their implementation in healthcare contexts, particularly in tasks such as ICD coding which require significant digit-level precision. [11]'s findings suggest that the GPT-4 LLM predicts an excessive number of ICD codes for medical coding tasks, leading to high recalls but low precisions. To tackle this challenge, they introduce LLM-codex, a two-stage approach to predict ICD codes that first generates evidence proposals using an LLM and then employ an LSTM-based verification stage. Our method is different from these methods, as we use the multi-agent system and inject external knowledge of code description to the LLM without any pre-training or fine-tuning.

**Multi-agent Systems** With LLMs showing remarkable promise in achieving human-like intelligence , it also increased the interest in context-aware multi-agent systems (MAS). These are systems of multiple agents that can act, perceive, and communicate in a shared environment, and that can adapt their knowledge to the perceived context and optimally solve assigned tasks. MAS have various applications in domains such as autonomous driving, disaster relief, utility management, supply chain management, human-AI interaction, cybersecurity, and other complex problems[35]. MAS has long been studied in healthcare applications for improving the accuracy of diagnoses and treatment plans and hence leading to better patient outcomes [36,37]. [38] employs task-specific fine tuning and self-play to enhance conversational AI's ability in diagnosis. In this work, we apply MAS to the complex task of ICD coding and explore different ways of enhancing its performances.

**Method**

In this section, we describe the details of our multi-agent method and its components, the roles and behaviors of each agent and how they interact with each other.

The multi-agent system consisted of five participants as stated above: a patient, a physician, a coder, a reviewer, and an adjustor. It mimics the coding procedure in a large healthcare system.

1. The Patient agent is the one who receives the medical service and the corresponding ICD codes at the healthcare facilities. They can review the codes and appeal to the adjustor agent if they find any errors or overcharges.

2. The Physician agent is the one who provides the service, documents observations, interventions, and outcomes in the discharge summary. They also check the codes generated by the coder agent and the reviewer agent, and can raise an issue to the adjustor agent if they find any errors or discrepancies.

3. The Coder agent is the one who generates the ICD codes based on the clinical notes written by the physician agent. They also try to optimize the code coverage for the health care facilities by choosing the most appropriate codes.

4. The Reviewer agent is the one who verifies and modifies the ICD codes produced by the coder agent. They play a vital role as their reviewed codes will be scrutinized by the patients and the physicians. They aim to minimize the patient's complaints and ensure the accuracy and quality of the codes.

5. The Adjustor agent is the final arbiter of the ICD codes when the patient or the physician disagrees with the reviewer agent. They are only invoked in case of such conflicts and they can adjust the codes as needed.

After defining the roles of the five agents, we designed two work flow modes for them, called Multi-Agent Coding I (MAC-I) and Multi-Agent Coding II (MAC-II). Figure I shows the work framework of MAC-I. The process is as follows:

1. The coder receives the discharge summary and generates codes.

2. The reviewer gets the discharge summary and codes and makes adjustments if needed.

3. The patient and physician review the discharge summary and the revised codes.

4. The patient and/or physician can appeal to the adjustor if they object to the codes; otherwise, the codes are final.

5. The adjustor checks the codes and summary and assigns the final codes.

Hallucination has been studied as a prevalent problem for LLMs in automatic ICD coding [5]. To mitigate this challenge, we propose MAC-II, which leverages the clinical knowledge embedded in the structure of EHRs, specifically the SOAP (subjective, objective, assessment, and plan) format. Physicians write notes following the SOAP structure, where the assessment and plan sections can be inferred from the subjective and objective sections [39]. Our method consists of two steps: first, we use an LLM to convert the discharge summary notes into the SOAP format; second, we apply our agents to perform the ICD coding task based on the following workflows.

1. The physician generates the assessment and plan sections based on the subjective and objective sections.
2. The physician compares the generated assessment and plan with the original gold standard assessment and plan section to check for accuracy and completeness, identify inconsistencies and generate the ICD codes.
3. The patient and physician review the codes and report any discrepancies or disagreements to the adjustor.
4. The adjustor reviews the discharge summary and the assigned ICD codes, and resolves any conflicts or errors for the final decision.

In MAC-II, the physician takes over the coder's role. The physician generates the assessment and plan sections based on the subjective and objective sections, and compares them with the original assessment and plan sections to conduct self-correction. Then, the physician assigns the ICD codes.

For both methods, the agents were equipped with the following strategy and external knowledge to enhance their functions for better ICD coding accuracy:

1. Confrontation Strategy: The coder (or the physician in method II) is instructed to assign as many codes as possible to reflect all the services rendered and also optimize the payment for the health care facilities, while the patient is encouraged to review the codes to prevent being overbilled.
2. External Knowledge: The agents use the candidate codes and their code descriptions as references to enhance their ICD coding performance. This approach is widely used by previous work on ICD coding [11].

| Role | Output |
| --- | --- |
| Coder | "code": "401.9","explanation": "Hypertension" <br> "code":"569.81","explanation":"Discharge Diagnosis: Peptic ulcer" ... |
| Reviewer | "code": "401.9","explanation": "Hypertension" <br> "code":"569.81","explanation":"Discharge Diagnosis: Peptic ulcer" ... |
| Physician | "code": "401.9","explanation": "Patient has a history of hypertension and was continued on losartan during hospitalization." <br> "code":"569.81","explanation":"Discharge Diagnosis: Peptic ulcer"... |
| Patient | "code": "401.9","explanation": "No evidence of hypertension found in the text." <br> "code":"569.81","explanation":"Discharge Diagnosis: Peptic ulcer"... |
| Adjustor | "code": "401.9", "explanation": "Mr. [**Known last name 85439**] was initially managed in [**Location (un) 20338**],FL for a CHF exacerbation with milrinone and dopamine. Diuresis was deferred due to hypotension." <br> "code":"569.81","explanation":"Discharge Diagnosis: Peptic ulcer" ... |

Figure 2: A snippet of the output of each agent in MAC-I, prepossessed for better readability. The patient complained about the hypertension ICD code, and the adjustor decided this code is related based on the healthcare note.

**Experiments**

In this section, we evaluate our methods on the MIMIC-III dataset and compare it with the state-of-the-art methods. We use the GPT-* API in Azure OpenAI Service which is recommended by physionet to process MIMIC dataset [1].

**Dataset and Metrics**

1. MIMIC-III top 50: We use the public MIMIC-III dataset of ICU discharge summaries with expert ICD codes. As in [19], we only keep instances with one or more of the top 50 ICD codes, resulting in 1,729 test instances.

2. MIMIC-III rare: Following [11], we create the rare code dataset from MIMIC-III to test our method's efficacy, underscoring their significance alongside common codes.

3. MDACE Profee: For explainability evaluation, we use the MDACE dataset [40], which provides code evidence for a subset of the MIMIC-III clinical records. Professional medical coders annotate a short text span for each ICD code, indicating the rationale for the code assignment. The MDACE Profee datasets are annotated following the professional fee billing rules. We use the sentences that contain the annotated text spans as the evidence for evaluation. The evaluation dataset consists of 172 sentence-ICD pairs.

For comparison, we followed previous work and report a variety of metrics, focusing on the precision, recall, micro-averaged and macro-averaged F1. Micro-averaged values are calculated by treating each (text, code) pair as a separate prediction. Macro-averaged values, are calculated by averaging metrics computed per-label thus plays much more emphasis on rare label prediction.

4.2 Comparison with State-of-the-Art Methods

We evaluated our proposed multi-agent coding method MAC-I and MAC-II against the following baselines:

- **CAML** [19] employs a convolutional attention network that adapts to different segments of the clinical note for each code.

- **MSMN** [41] applies code description synonyms with multi-head attention and outperforms existing methods on the MIMIC-III common task.

- **EffectiveCAN** with supervised attention [40] uses a convolutional attention network to learn from document-level labels and evidence annotations with supervised attention. Their evidence annotations are from clinical coding experts.

- **LLM-codex**, [11] a two-stage approach to predict ICD codes that first generates evidence proposals using an LLM and then employs an LSTM-based verification stage.

- **Zero-shot CoT prompting** (CoT) [42] Inspired by the chain-of-thought method [43], this zero-shot prompting approach adds "Let's think step by step" to the input query, making the model think more carefully and gradually before solving the problem or task without any examples provided. We evaluated using GPT-4 and GPT4o respectively.

---
[1] https://physionet.org/news/post/gpt-responsible-use

- **Self-consistency with CoT (CoT-SC)** [44]refines LLMs' problem-solving by generating multiple reasoning paths and selecting the most consistent answer, boosting performance on complex tasks. We evaluated using GPT-4 and GPT4o respectively.

- **LLM-designed agent:** prompt an LLM to automatically generate and assign agent roles for the ICD coding task, enabling them to work in synergy or contention to enhance task performance. We evaluated using GPT-4 and GPT-4o respectively. GPT-4 automatically designed seven roles respectively as Preprocessing Agent, Natural Language Processing (NLP) Agent, Medical Entity Mapping Agent, ICD Coding Agent, Validation and Quality Assurance Agent, User Interface (UI) Agent and Feedback Learning Agent. GPT-4o automatically designed six roles respectively as data extraction agent, preprocessing agent, medical term mapping agent, ICD code assignment agent, quality assurance agent, and reporting agent.

**Results**

5.1 Results on MIMIC-III top 50 Codes

Table 1 shows the results of MAC methods and the state-of-the-art methods on the test set. We can see that our proposed MAC-II method outperforms the state-of-the-art methods on Macro-F1 with a score of 0.748. The multi-agent systems have shown significant improvements in performing such complex tasks, comparing to CoT style reasoning and LLM-designed agents. Both MAC-I and MAC-II achieved a Micro-F1 score above 0.58, which is lower than the two-stage LLM-codex method but higher than all others. This suggests that our methods can assign more accurate and complete codes to the clinical notes without model training/tuning compared with the baselines. On the other hand, the well-designed task-specific fine-tuned models show their superiority in data-rich scenarios, as expected. Also, the results show that the MAC-II in general outperforms MAC-I, which demonstrates the effectiveness in utilizing the SOAP structure of clinical notes.

Table 1: Results on the test set, MIMIC-III top 50

| Method | | Precision | Recall | Micro-F1 | Macro-F1 |
|---|---|---|---|---|---|
| CAML | | —— | —— | 0.364 | 0.258 |
| MSMN | | —— | —— | 0.561 | 0.489 |
| EffectiveCAN | | —— | —— | 0.556 | 0.434 |
| LLM-codex | | —— | —— | **0.611** | 0.468 |
| CoT | GPT-4 (1 call) | 0.124 | 0.210 | 0.156 | 0.419 |
| | GPT-4o (1 call) | 0.109 | 0.183 | 0.136 | 0.403 |
| CoT-SC | GPT-4 (5 calls) | **0.569** | 0.560 | 0.564 | 0.726 |
| | GPT-4o (5 calls) | 0.394 | 0.627 | 0.484 | 0.624 |
| LLM-designed agent | GPT-4 (7 calls) | 0.56 | 0.475 | 0.514 | 0.743 |
| | GPT-4o ( 6 calls) | 0.432 | 0.542 | 0.481 | 0.667 |
| MAC-I | GPT-4 (5 calls) | 0.535 | 0.644 | 0.585 | 0.693 |
| | GPT-4o (5 calls) | 0.474 | **0.763** | 0.584 | 0.670 |
| MAC-II | GPT-4 (5 calls) | 0.551 | 0.633 | **0.589** | **0.748** |
| | GPT-4o (5 calls) | 0.502 | 0.703 | 0.586 | 0.725 |

## 5.2 Results on MIMIC-III rare Codes

For the rare codes, we achieve significant improvements on all metrics. Our proposed MAC-II method obtains a Micro-F1 of 0.376 (GPT-4), outperforming all baselines, even the well-designed task-specific fine-tuned ones. This demonstrates the advantages of our method for classification with scarce data. It also implies that fine-tuned systems like MSMN, LLM-codex, etc., can benefit from more data of these uncommon categories.

Table 2: Results on the test set, MIMIC-III rare.

| Method | | Precision | Recall | Micro-F1 | Macro-F1 |
|---|---|---|---|---|---|
| CAML | | —— | —— | 0.083 | 0.072 |
| MSMN | | —— | —— | 0.173 | 0.169 |
| LLM-codex | | —— | —— | 0.302 | 0.279 |
| CoT | GPT-4 | 0.164 | 0.402 | 0.233 | 0.030 |
| | GPT-4o | 0.127 | 0.193 | 0.153 | 0.042 |
| CoT-SC | GPT-4 | 0.260 | 0.455 | 0.331 | 0.659 |
| | GPT-4o | 0.127 | 0.364 | 0.189 | 0.579 |
| LLM-designed agent | GPT-4 | 0.263 | 0.455 | 0.333 | 0.660 |
| | GPT-4o | 0.115 | 0.333 | 0.171 | 0.700 |
| MAC-I | GPT-4 | 0.316 | 0.436 | 0.366 | 0.704 |
| | GPT-4o | 0.194 | **0.636** | 0.298 | **0.881** |
| MAC-II | GPT-4 | **0.322** | 0.452 | **0.376** | 0.715 |
| | GPT-4o | 0.199 | 0.583 | 0.297 | 0.729 |

## 5.3 Results on MDACE Profee

Table 3 presents the results on the MDACE Profee dataset. The proposed methods achieve on par results with fine-tuned systems in terms of F-1 score. However, we observe that the proposed methods have higher precisions and lower recalls than the fine-tuned models. This could be due to the prompts that we use to guide the model to be accurate. It could also be related to the model's inherent characteristics. The underlying reason requires further investigation.

Table 3: Results on the test set, MDACE Profee

| Method | | Precision | Recall | F1 |
|---|---|---|---|---|
| EffectiveCAN | | 0.408 | 0.806 | 0.542 |
| LLM-codex | | 0.608 | **0.861** | **0.713** |
| MAC-I | GPT-4 | 0.875 | 0.601 | 0.712 |
| | GPT-4o | **0.882** | 0.583 | 0.709 |
| MAC-II | GPT-4 | 0.836 | 0.605 | 0.702 |
| | GPT-4o | 0.855 | 0.603 | 0.707 |

## 5.4 Ablation Study

Our initial ablation study examined the impact of the confrontation strategy and the integration of external knowledge into the system. The results in Table 4 indicate that both the confrontation strategy and external knowledge enhance ICD coding performance. Notably, the confrontation strategy has a more significant effect than external knowledge when omitted from the prompts. Our analysis shows that the confrontation strategy aids in assigning more codes, primarily by increasing recall. This suggests that these strategies can be further explored to improve the performance of MAC methods. Considering that not all ICD codes hold equal importance, we can incorporate knowledge such as code billing costs into the system for better performance for future work.

We conducted additional experiments to assess the impact of each agent. Using the MAC-I on MIMIC-III rare as an example, our findings, as shown below, indicate that certain agents significantly affect performance. Specifically, we identified that the Reviewer and Physician agents play pivotal roles. Interestingly, we observed that including the "patient" agent led to a drop in the recall metric but an increase in precision. This may suggest that patients are concerned about being overcharged, or it might reflect the reality of the situation. Additional human evaluation by domain experts is needed to verify this.

Table 4: Ablation results on the test set with MAC-II, MIMIC-III rare (GPT-4)

| Method | Precision | Recall | Micro-F1 | Macro-F1 |
|---|---|---|---|---|
| MAC-II without Confrontation Strategy | 0.311 | 0.430 | 0.361 | 0.703 |
| MAC-II without External Knowledge | 0.314 | 0.445 | 0.368 | 0.706 |
| MAC-II | 0.322 | 0.452 | 0.376 | 0.715 |

Table 5: Ablation results on the test set with MAC-I, MIMIC-III rare on (GPT-4)

| | Precision | Recall | Micro-F1 | Macro-F1 |
|---|---|---|---|---|
| Coder | 0.174 | 0.441 | 0.250 | 0.607 |
| Coder+Reviewer | 0.196 | 0.468 | 0.276 | 0.644 |
| Coder+Reviewer+Patient | 0.258 | 0.407 | 0.316 | 0.605 |
| Coder+Reviewer+Physician | 0.320 | 0.420 | 0.363 | 0.701 |
| Coder+Reviewer+Physician+Patient+Adjustor | 0.316 | 0.436 | 0.366 | 0.704 |

5.5 Applicability

To in line with prior studies such as LLM-Codex, CAML, etc. in the literature, our work focuses on the top-50 and rare ICD-9 codes, it also allows for the embedding of external knowledge within the context limit. However, we notice that such setting may not fully represent real-world scenarios where a vast array of ICD-9/10 codes could be applicable. To better align with real-world applications, we also conducted experiments excluding the top-50 and rare ICD-9 codes constraint to provide a clearer picture of the method's performance in a more generalized setting. The results are shown in Table 6. As is shown, the proposed MAC method can achieve a Micro F1 of above 0.33 and Marco F1 of above 0.68, beating the strong baseline of LLM-designed agents. This proves the effectiveness of the designed agents.

Table 6: ICD Coding Results Excluding Top-50/Rare ICD-9 Constraints for MAC-II (GPT-4)

| Method | Precision | Recall | Micro F1 | Macro-F1 |
|---|---|---|---|---|
| LLM-designed agent | 0.440 | 0.247 | 0.316 | 0.629 |
| MAC-I | 0.505 | 0.251 | 0.335 | 0.682 |
| MAC-II | 0.511 | 0.263 | 0.347 | 0.702 |

**Discussion**

Although LLMs have been recognized for their superior performance in various tasks and domains, their efficacy as medical coders have been previously questioned [5,11]. In this study, we present an advanced multi-agent coding framework that leverages LLMs. We evaluated our approach using real-world clinical narratives/notes from the EHR. Our experiments show that our method achieves competitive performance in ICD coding and outperforms existing methods in rare code mapping and interpretability.

Contrasted with LLM-generated multi-agent frameworks, our expertly crafted system outperforms with less LLM computations, it more accurately reflects the intricate coding procedures found in healthcare settings. It adeptly models the nuanced interactions between patients, physicians, and coders. This innovative and pragmatic strategy, previously unexamined in academic research, allows for the assignment of more precise and comprehensive ICD codes to clinical documentation, effectively managing the inherent uncertainties and complexities of medical coding.

Our method provides informative and relevant explanations for each code, justifying the coding decisions. This enhances the explainability and transparency of our method, helping users understand and verify the codes, and make complaints and corrections if necessary. Our method explores the strengths and possibilities of LLM-based models for ICD coding. Previous work that used prompts to guide LLMs for ICD coding suffered from low precision and hallucination issues. Our multi-agent method, which does not require any training or fine-tuning, achieves competitive performance with state-of-the-art methods. This demonstrates the potential of LLMs for ICD coding and other healthcare tasks that we plan to investigate further.

Our method has several implications and applications for ICD coding and other clinical NLP tasks. It can improve the quality and efficiency of ICD coding, which is essential for billing, epidemiology, and quality improvement in the healthcare domain. The designed system mimics real-world healthcare systems. From the results in Table 5, our analysis reveals that both the Reviewer and Adjuster roles significantly enhance final performance metrics. Conversely, the Patient role appears to negatively influence the F1 scores, suggesting potential issues with over-billing. The observed discrepancy between the Patient and Physician roles underscores the importance of effective communication. These results are for reference, and real-world validation is necessary to confirm these findings.

Our method can also extend to other clinical NLP tasks, such as diagnosis prediction, treatment recommendation, and adverse event detection, which share similar characteristics and objectives with ICD coding. By adapting the agents and prompts, our method can broaden its scope and applicability.

**Limitations**

Our method also has some limitations and drawbacks, which we acknowledge and plan to address in the future work. Although our devised framework has significantly enhanced the performance of LLMs, the ICD coding task itself demands exceptionally high precision. This is crucial because even minor errors can lead to significant consequences in billing, epidemiology, and patient care. Therefore, while our system shows promise, it still requires further refinement and rigorous testing before it can be implemented in real-world clinical settings. We need to focus on improving the accuracy and reliability of our method to ensure it meets the stringent standards required for practical application. Additionally, ongoing evaluation and feedback from healthcare professionals will be essential to fine-tune the system and address any potential issues that arise during its deployment.

Furthermore, our method relies on LLM-based models, which are subject to ethical and social issues and may generate results that affect the decisions, actions, or outcomes of users or patients. We use the GPT-4 and GPT-4o APIs in Azure OpenAI Service, as recommended by PhysioNet, to process the MIMIC dataset and ensure data security. However, attention must be paid to these concerns, and open-source LLM models are recommended for deployment in healthcare platforms.

A    Appendix

A.1    Implementation Details

We use GPT4-8k version (Achiam et al., 2023) as the LLM for this study. We access it securely via the Azure OpenAI API with the responsible use requirement. We truncate the EHRs to fit the 8k token limit and sample with a temperature of 0.1. We also set and test the number of candidate codes Nc to 50, which could change for different applications.

A.2    Prompts

SOAP format Prompt:
You are an assistant who convert the EHR note into SOAP format. Please format the note in the SOAP (subjective, objective, assessment, plan) format. The output is a valid JSON as is shown below.

```
{
    "Subjective": content,
    "Objective": content,
    "Assessment": content,
    "Plan": content
}
```

Patient Prompt: You are a patient who receieved treatment at the hospital. You cooperate fully with the health care system to receive the best service possible. You also check the ICD-9 codes to avoid being overbilled. You check all assigned ICD-9 codes and explain the reasons for each code.

Physician Prompt: v1
(in MAC-I):
You are a physician who treats patients. You strive to provide the best service to each patient. You document your findings, interventions and results in the discharge summary note. You check all assigned ICD-9 codes and explain the reasons for each code.

v2 (in MAC-II, AP generation):
You are a physician who treats patients. You strive to provide the best service to each patient. Based on the Subjective and Objective, you will generate the assessment and plan for the EHR note.

v3 (in Method MAC-II, code assignment):
You are a physician who treats patients. Please check the generated assessment and plan against the gold standard assessment and plan. Please pay attention to the inconsistencies You assign ICD-9 codes to the note. You assign as many as possible ICD-9 codes and explain the reasons for each code.

---

Coder prompt:
You are an ICD-9 coder. You assign ICD-9 codes to the discharge summary based on the clinical care that the patients received. You cite the discharge summary as evidence when needed. You assign as many as possible ICD-9 codes and explain the reasons for each code.

---

Reviewer Prompt:
You are a reviewer. You will check the ICD-9 codes assigned by the coder. You can use the ICD-9 dictionary for guidance. Your role is to ensure that the assigned ICD-9 codes are correct. You assign all possible ICD-9 codes and explain the reasons for each code.

---

Adjustor Prompt:
When a patient or a physician has different thoughts about the ICD-9 codes, you will review the discharge summary and the ICD codes assigned by the coder and checked by the reviewer. You can add, remove the assigned codes to make them accurate. You can consult the ICD-9 dictionary for assistance. Your duty is to ensure that the assigned ICD-9 codes are valid and exact. You assign all possible ICD-9 codes and explain the reasons for each code.

---

**LLM-designed Agents Prompt**

You are a useful assistant. You will create several agents to work together to assign ICD codes to patient discharge summaries. Please generate all the necessary agents and describe their responsibility.

**GPT-4 designed roles**
**Preprocessing Agent**: This agent is responsible for cleaning and standardizing the patient discharge summaries. It removes irrelevant information, normalizes medical terminology, and structures the text to be easily analyzed by subsequent agents.
**Natural Language Processing (NLP) Agent**: This agent applies natural language processing techniques to understand the context, extract medical entities (symptoms, diagnoses, treatments, etc.), and interpret the meanings of sentences within the processed discharge summaries.
**Medical Entity Mapping Agent**: Following entity extraction, this agent is tasked with mapping extracted medical entities to standardized codes using databases like UMLS (Unified Medical Language System). It utilizes synonyms and related concepts to ensure accurate mapping.
**ICD Coding Agent**: Utilizing the mapped medical entities, the ICD Coding Agent assigns the appropriate International Classification of Diseases (ICD) codes to the patient's conditions and procedures noted in the discharge summaries. This includes identifying primary and secondary diagnoses as well as procedure codes relevant to the patient's stay.
**Validation and Quality Assurance Agent**: This agent checks the assigned ICD codes for accuracy and completeness. It reviews code assignments against clinical guidelines and coding standards to ensure that all codes are appropriately justified by the text of the discharge summary.

**User Interface (UI) Agent**: Serves as the interface between the coding system and the human coders or healthcare professionals. This agent allows users to review, edit, and confirm the automatically assigned ICD codes, ensuring a final layer of verification and accuracy.

**Feedback Learning Agent**: Learns from corrections and adjustments made by human coders to improve the accuracy of automatic ICD coding over time. This agent adapts its algorithms based on feedback to minimize errors and refine the performance of the coding process.

**GPT-4o designed roles**

**Data Extraction Agent:** Extracts relevant medical information and key terms from patient discharge summaries.

**Preprocessing Agent:** Cleans and preprocesses the extracted information to remove any irrelevant or redundant data, and prepares it for analysis.

**Medical Term Mapping:** Maps medical terminologies and symptoms mentioned in the preprocessed data to corresponding standardized terms in medical dictionaries such as SNOMED CT.

**ICD Code Assignment Agent:** Assigns appropriate ICD codes to the mapped medical terms based on standardized guidelines.

**Quality Assurance Agent:** Reviews the assigned ICD codes for accuracy and consistency, ensuring they adhere to current medical coding standards.

**Reporting Agent:** Generates a comprehensive report that includes the original discharge summary, extracted information, mapped medical terms, assigned ICD codes, and quality assurance notes.

External Knowledge:

Please only use ICD-9 codes that are listed below:

Candidate Codes:

401.9 : Unspecified essential hypertension

38.93 : Venous catheterization, not elsewhere classified

428.0 : Congestive heart failure, unspecified

427.31 : Atrial fibrillation

414.01 : Coronary atherosclerosis of native coronary artery

96.04 : Insertion of endotracheal tube

96.6 : Enteral infusion of concentrated nutritional substances

…